\documentclass[sigconf]{acmart}
\AtBeginDocument{%
  \providecommand\BibTeX{{%
    \normalfont B\kern-0.5em{\scshape i\kern-0.25em b}\kern-0.8em\TeX}}}

\author{Thai Le} 
\affiliation{%
  \institution{The Pennsylvania State University}
}
\email{thaile@psu.edu}

\author{Suhang Wang}
\affiliation{%
  \institution{The Pennsylvania State University}
}
\email{szw494@psu.edu}

\author{Dongwon Lee}
\affiliation{%
  \institution{The Pennsylvania State University}
}
\email{dongwon@psu.edu}

\setcopyright{acmcopyright}
\copyrightyear{2020}
\acmYear{2020}
\acmDOI{10.1145/3394486.3403066}

\usepackage{hyperref}
\hypersetup{
  pdfinfo={
  pdfproducer={},
  Title={},
  Subject={},
  Author={},
  }
}

\usepackage{enumitem}
\usepackage{url}
\usepackage{caption}
\usepackage{algorithm}
\usepackage[noend]{algcompatible}
\usepackage{tabularx}
\usepackage{amsmath,amsfonts}
 
\usepackage{amssymb}
\usepackage{subfiles}
\usepackage{stfloats}
\usepackage{textcomp}
\usepackage{xcolor}
\usepackage{amsthm}
\usepackage{subcaption}
\usepackage{array}
\usepackage{booktabs}
\usepackage{multirow}
\usepackage{amsmath}
\usepackage{mathtools}
\usepackage{varwidth}

\newtheorem{definition}{\textsc{Definition}}

\newcolumntype{H}{>{\setbox0=\hbox\bgroup}c<{\egroup}@{}}

\newcolumntype{$}{>{\global\let\currentrowstyle\relax}}
\newcolumntype{^}{>{\currentrowstyle}}

\newcommand{\beginsupplement}{%
        \setcounter{table}{0}
        \renewcommand{\thetable}{Sup\arabic{table}}%
        \setcounter{figure}{0}
        \renewcommand{\thefigure}{Sup\arabic{figure}}%
     }
     
\newcommand\bigforall{\mbox{\Large $\mathsurround0pt\forall$}} 
     
\newcolumntype{H}{>{\setbox0=\hbox\bgroup}c<{\egroup}@{}}

\DeclareMathAlphabet\mathbfcal{OMS}{cmsy}{b}{n}

\usepackage{bbm}

\usepackage[compact]{titlesec}

\copyrightyear{2020} 
\acmYear{2020} 
\setcopyright{acmcopyright}\acmConference[KDD '20]{Proceedings of the 26th ACM SIGKDD Conference on Knowledge Discovery and Data Mining}{August 23--27, 2020}{Virtual Event, CA, USA}
\acmBooktitle{Proceedings of the 26th ACM SIGKDD Conference on Knowledge Discovery and Data Mining (KDD '20), August 23--27, 2020, Virtual Event, CA, USA}
\acmPrice{15.00}
\acmDOI{10.1145/3394486.3403066}
\acmISBN{978-1-4503-7998-4/20/08}

\begin{document}
\fancyhead{}
\newcommand{\mymethod}{{GRACE}}
\setlength\abovedisplayskip{2pt}
\setlength\belowdisplayskip{2pt}
\title{
{\sf \mymethod}: Generating Concise and Informative Contrastive Sample to Explain Neural Network Model's Prediction}

\begin{abstract}
Despite the recent development in the topic of explainable AI/ML for image and text data, the majority of current solutions
are not suitable to explain the prediction of neural network models when the datasets are tabular and their features are in high-dimensional vectorized formats. To mitigate this limitation, therefore, we borrow two notable ideas (i.e., ``explanation by intervention" from causality and ``explanation are contrastive" from philosophy) and propose a novel solution, named as  {\mymethod}, that better explains neural network models' predictions for tabular datasets. In particular, given a model's prediction as label $X$, {\mymethod} intervenes and generates a minimally-modified contrastive sample to be classified as $Y$, with an intuitive textual explanation, answering the question of  ``Why $X$ rather than $Y$?" We carry out comprehensive experiments using eleven public datasets of different scales and domains (e.g., \# of features ranges from 5 to 216) and compare {\mymethod} with competing baselines on different measures: \textit{fidelity}, \textit{conciseness}, \textit{info-gain}, and \textit{influence}. The user-studies show that our generated explanation is not only more intuitive and easy-to-understand but also facilitates end-users to make as much as 60\% more accurate post-explanation decisions than that of \textsc{Lime}. We release the source code of \textsc{\mymethod} at: {\tt  https://github.com/lethaiq/GRACE\_KDD20}

\end{abstract}




\maketitle

\vspace*{-5pt}
\section{INTRODUCTION}

\begin{table}[t]
\small
\vskip 0.2cm
\begin{tabular}{lrrrrr}
\toprule
{Feat} & freq\_now &  freq\_credit &  freq\_!!! &  freq\_! &  class \\
\hline
$\pmb{x}_1$ & 0.1 & 0.0 & 0.0 & 0.0 & Ham\\
$\mathbf{\Tilde{x}_1}$ & 0.1 & 0.0 & \textbf{0.3} & \textbf{0.453} & \textbf{Spam}\\
\end{tabular}
\smallskip
\begin{tabular}{lrrrrr}
\toprule
{Feat} &  freq\_you &  freq\_direct & avg\_longest\_capital  &  class \\
\hline
$\pmb{x}_2$ & 0.68 & 0.34 & 158.0 & Spam\\
$\mathbf{\Tilde{x}}_2$  & 0.68 & 0.34 & \textbf{1.0} & \textbf{Ham}\\
\bottomrule
\end{tabular}
\caption{Examples of original samples $\pmb{x}_i$ and contrastive samples $\tilde{\pmb{x}}_i$ on \textit{spam} dataset. $\tilde{x}_i$ only differs from $\pmb{x}_i$ on \textit{a few} features.
(unchanged features are randomly selected)}
\vskip -1.3cm
\label{tab:example_spam}
\end{table}

Tabular data is one of the most commonly used data formats. Even though tabular data receives far less attention than computer vision and NLP data in neural networks literature, recent efforts (e.g., \cite{shavitt2018regularization,arik2019tabnet,barz2019deep,marais2019deep}) have shown that neural networks, deep learning in particular, can also achieve superior performance on this type of data. Yet, there is still a lack of interpretability that results in the distrust of neural networks trained on general tabular data domains. This obstructs the wide adoption of such models in many high-stakes scenarios in which tabular data is prominent--e.g., healthcare~\cite{ravi2016deep,bejnordi2017diagnostic}, finance~\cite{fischer2018deep,dixon2019deep}, social science~\cite{kosinski2013private,zizzo2000bounded}, and cybersecurity~\cite{mahdavifar2019application,xu2017neural}. Moreover, the majority of explanation algorithms (e.g., ~\cite{zeiler2014visualizing,simonyan2013deep,petsiuk2018rise,li2015visualizing,karpathy2015visualizing,lime,chu2018exact}) are designed for models trained on images or texts, while insufficient efforts have been made to explain the prediction results of neural models that take tabular data formats as input. Furthermore, most of the previous explanation approaches
are geared for professional users such as ML researchers and developers
rather than lay users and ML consumers.
This situation calls for a novel approach to provide end-users with the intuitive explanation of neural networks trained on tabular data. However, developing such an approach poses several challenges.
\vspace{0.05in} \\
\noindent{\bf Challenges.} 
\textit{First}, tabular data  used in neural network models sometimes have high-dimensional inter-correlated features. Therefore, presenting feature importance scores for top-$k$ or all features (e.g., \cite{lime}) can induce both information overload and redundancy, causing
confusion to end-users.
In fact, for a data instance, a complex model can focus on just a few key features in making its prediction. To illustrate, Table \ref{tab:example_spam} shows that for two emails $\pmb{x}_1$ and $\pmb{x}_2$, the model can focus on two different sets of features, $freq\_!$, $freq\_!!!$ or $avg\_longest\_capital$, respectively, to predict if an email is a spam or ham. While explanation constructed \textbf{only} from these features is much more concise, providing both $freq\_!$ and $freq\_!!!$ (frequency of ``!" and ``!!!" within an email content) in the first example produces redundancy. In this case, we also want to replace $freq\_!$ with another key feature to make the explanation more informative. Thus, we need to find a subset of instance-dependent key features that are both {\em concise} and {\em informative} to explain the model's prediction. 

\textit{Second}, for images or texts, highlighting a patch of an image (e.g., \cite{zeiler2014visualizing,li2015visualizing,chu2018exact}) or a phrase of a sentence (e.g., \cite{karpathy2015visualizing}) usually gives a clear  understanding of what a model is focusing on and why a model gives such prediction.
However, in tabular data, such visualization does not provide much insight into the chosen model. For instance, in the second example in Table~\ref{tab:example_spam}, the model predicts $\pmb{x}_2$ as spam and the important feature used by the model is $avg\_longest\_capital$. However, simply providing this feature to end-users does not give an easy-to-understand explanation. Since we often justify our decision verbally \cite{lipton2018mythos}, in this case, an explanation written in text can help end-users understand the prediction better.

\textit{Third}, approximating the decision boundaries does not necessarily provide a clear understanding on the decision-making of a model to end-users, who usually lack ML background. Instead, such lay users are usually more interested in the \textbf{contrastive explanation}, i.e., why $X$ rather than $Y$. For example, Table \ref{tab:example_spam} shows in the second example that {\ttfamily "had $avg\_longest\_capital$ (i.e., the average length of the longest capitalized words) been about 150 characters shorter, the email would have been classified as \textbf{ham rather than spam}"}. Hence, we need to come up with a new explanation model such as \textit{contrastive explanation} to better explain a model's prediction to lay end-users.
\vspace{0.05in} \\
\noindent
{\bf Overview.} To sum up, the effort towards generating an explanation that is easy for end-users to understand is challenging, yet also in great demand. Therefore, we propose a novel algorithm, {\mymethod} (\textit{\underline{G}ene\underline{RA}ting \underline{C}onstrastive sampl\underline{E}s}), which generates and provides end-users with intuitive and informative explanations for neural networks trained on general tabular data. Inspired from Database (DB) literature \cite{wu2013scorpion,meliou2014causality,roy2014formal}, {\sc \mymethod} borrows the idea of ``explanation by intervention" from causality \cite{lewis2013contrastives,silverstein2000scalable} to come up with contrastive explanation--i.e., why a prediction is classified as $X$ rather than $Y$. Specifically, for each prediction instance, {\sc \mymethod} generates an explainable sample and its contrastive label by selecting and modifying a few instance-dependent key features under both \textit{fidelity}, \textit{conciseness} and \textit{informativeness} constraints. Then, {\sc \mymethod} aims to provide a friendly text explanation of \textit{why $X$ rather than $Y$} based on the newly generated sample. 
The main contributions of the paper are:

\begin{itemize}[leftmargin=\dimexpr\parindent+0.1\labelwidth\relax]
    \item We introduce an explanation concept for ML by marrying ``contrastive explanation" and ``explanation by intervention", then extend it to a novel problem of generating contrastive sample to explain why a neural network model predicts $X$ rather than $Y$ for data instances of tabular format;
    \item We develop a novel framework, {\mymethod}, which finds key features of a sample, generates contrastive sample based on these features, and provides an explanation text on why the given model predicts $X$ rather than $Y$ with the generated sample; and
    \item We conduct extensive experiments using eleven real-world datasets to demonstrate the quality of generated contrastive samples and the effectiveness of the final explanation. Our user-studies show that our generated explanation texts are more intuitive and easy-to-understand, and enables lay users to make as much as 60\% more accurate post-explanation decisions than that of \textsc{Lime}.
\end{itemize}

\section{THE EXPLANATION MODEL}
\subsection{Contrastive Explanation}
Understanding the answer to the question "Why?" is crucial in many practical settings, e.g., in determining why a patient is diagnosed as benign, why a banking customer should be approved for a housing loan, etc. The answers to these "Why?" questions can be really answered by studying causality, which depicts the relationship between an event and an outcome. The event is a cause if the outcome is the consequence of the event (\cite{meliou2014causality, silverstein2000scalable}). However, 
causality can only be established under a controlled environment, in which one alters a single input while keeping others constant, and observes the change of the output. Bringing causality into data-based studies such as DB or ML is a very challenging task since causality cannot be achieved by using data alone (\cite{meliou2014causality}). As the first step to understand causality in data-intensive applications, DB and ML researchers have tried to lower the bar of explanation, aiming to find the subset of variables that are best correlated with the output. Specifically, DB literature aims to provide explanations for a complex query's outputs given all tuples stored in a database, while ML researcher is keen on explaining the predictions of learned, complex models.

\subsection{Explanation by Intervention}
By borrowing the notion of \textit{intervention} from causality literature, in particular, DB researchers have come up with a practical way of explaining the results of a database query by searching for an explainable predicate $\mathcal{P}$. Specifically, $\mathcal{P}$ is an explanation of outputs $\mathbf{X}$ if the removal of tuples satisfying predicate $\mathcal{P}$ also changes $\mathbf{X}$ while keeping other tuples unchanged (\cite{meliou2014causality,wu2013scorpion,roy2014formal}). Similarly, by utilizing the same perspective, we want to formulate a definition of \textit{explanation by intervention} for ML models at instance-level as follows.

\vspace{-3pt}
\noindent\begin{definition}[Contrastive Explanation (in ML) by Intervention]
    \label{def:intervention}
    A predicate $\mathcal{P}$ of subset of features is an explanation of a prediction outcome $\mathbf{X}$, if changes of features satisfying the predicate $\mathcal{P}$ also \textit{changes} the prediction outcome to $\mathbf{Y}$($\not=\mathbf{X}$), \textbf{while \textit{keeping other features unchanged}}.
\end{definition}

For example, possible predicates to explain a spam detector are shown in Table \ref{tab:example_spam}. Particularly, predicate $\mathcal{P}_2: "\mathrm{avg\_longest\_capital=1.0}"$ explains why sample $\pmb{x}_2$ is classified as spam rather than ham. Given a prediction of a neural network model on an input, there will be possibly many predicates $\mathcal{P}$ satisfying Def. \ref{def:intervention}. Hence, it is necessary to have a measure to describe and compare how much influence predicate(s) $\mathcal{P}$ have on the final explanation. Following the related literature of explanation from the DB domain \cite{wu2013scorpion}, we also formally define a scoring function $\mathit{infl}_\lambda(\mathcal{P})$ as the measure on the influence of $\mathcal{P}$ on the explanation with a tolerance level $\lambda$ as.

\noindent\begin{definition}[Influence Scoring Function]
\begin{equation}
    \mathit{infl}_\lambda(\mathcal{P}) = \frac{\mathbbm{1}(\mathbf{Y} \not= \mathbf{X})}{(\textrm{Number of features in } \mathcal{P})^\lambda}
\end{equation}
where $\mathbbm{1}(\cdot)$ is an indicator function, $\mathbf{X}$ and $\mathbf{Y}$ are predicted labels before and after intervention, respectively.
\label{def:influencescore}
\end{definition}

\noindent{}The larger the score is, the more influential $\mathcal{P}$ has on the explanation. Hence, $\lambda = 0$ would imply infinite tolerance on the number of features in $\mathcal{P}$, $\lambda > 0$ would prefer a small size of $\mathcal{P}$ and $\lambda < 0$ would prefer a large size of $\mathcal{P}$. In practice, $\lambda > 0$ is preferable because a predicate $\mathcal{P}$ containing too many features would adversely affect the comprehension of the explanation. For example, $\mathit{infl}_1(\mathcal{P}_2) = 1.0$

\subsection{From Intervention to Generation}
Searching for $\mathcal{P}$ is a non-trivial problem. From Def. \ref{def:intervention}, we want to approach this problem from a generation perspective. Particularly, given an arbitrary sample classified as $\mathbf{X}$ by a neural network, we want to intervene and modify a small subset of its features to generate a new sample that crosses the decision boundary of the model to class $\mathbf{Y}$. This subset of features and their new values will result in a predicate $\mathcal{P}$. This newly generated sample will help answer the question ``Why X rather than Y?". To illustrate, Table \ref{tab:example_spam} shows that $\mathbf{\Tilde{x}_2}$ is  generated samples that correspond to predicate $\mathcal{P}_2:~"\mathrm{avg\_longest\_capital=1.0}"$. Using $\mathcal{P}_2$, we can generate an explanation text to present to the users such as {\ttfamily "Had the average length of the longest capitalized words been 1.0, the message would have been classified as {\bf ham rather than spam}"}.


\section{OBJECTIVE FUNCTION}\label{sec:problem}
Let $\mathit{f}(\cdot)$ be a neural network model that we aim to give instance-level explanation. Denote $\mathcal{X} \in \mathbb{R}^{N \times M} = \{x_1, x_2,..x_n\}$, $\mathcal{Y} = \{y_1, y_2,..y_n\}$ as the features and ground-truth labels of data on which $\mathit{f}(\pmb{x})$ is trained, where $N, M$ is the number of samples and features, respectively.
$\mathcal{X}^\mathit{i}$ and $\mathcal{X}^\mathit{j}$ are the $\mathit{i}$-th and $\mathit{j}$-th feature, respectively, in features set $\mathcal{X}$. $\pmb{x}^\mathit{i}$ and $\pmb{x}^\mathit{j}$ are the $\mathit{i}$-th and $\mathit{j}$-th feature  of $\pmb{x}$, respectively. First, we want to generate samples that are contrastive. We define such characteristic as follows. 

\begin{definition}[Contrastive Sample]
    Given an arbitrary $\pmb{x} \in \mathcal{X}$, $\mathcal{X} \in \mathbb{R}^{N \times M}$ and neural network model $\mathit{f}(\cdot)$, $\mathbf{\tilde{x}}$ is called contrastive or contrastive sample of  $\pmb{x}$ when:
    \begin{equation}
        \min_{\tilde{\pmb{x}}} dist(\pmb{x}, \tilde{\pmb{x}}) \quad s.t. \quad \mathrm{argmax}(f(\pmb{x})) \ne \mathrm{argmax}(f(\tilde{\pmb{x}}))   
        \label{cond1}
    \end{equation}
\end{definition}

\setlength{\fboxsep}{2pt}

Then, formally, we study the following problem:
\noindent
\begin{center}
\fbox{\parbox[t]{0.9\linewidth}{
\textbf{\textsc{Problem}}: Given $\pmb{x}$ and neural network model $\mathit{f}(\cdot)$, our goal is to generate new contrastive sample $\mathbf{\tilde{x}}$ to provide concise and informative explanation for the prediction $\mathit{f}(\pmb{x})$.
\label{problem}
}}
\end{center}

Existing works on adversarial example generation \cite{moosavi2016deepfool,mao2017least} usually define $dist(\tilde{\pmb{x}}, \pmb{x})$ as $\Vert\pmb{x} - \mathbf{\tilde{x}}\Vert_2^2$, which allows all features to be changed to generate $\tilde{\pmb{x}}$. Though such approaches can generate realistic labeled contrastive samples, they are not appropriate for generating instance-level explanation that are easy to understand because all features are changed.

Instead, we aim to generate $\mathbf{\tilde{x}}$ labeled $\mathbf{\tilde{y}}$ such that it is minimally different from original input $\pmb{x}$ in terms of only a \textit{few important features} instead of \textit{all features}.
Specifically, we desire to explain ``Why X rather than Y?" by presenting a \textit{concise} explanation in which only a few features are corrected, e.g., the first example of Table \ref{tab:example_spam} shows if only frequency of '!' is increased to 45.3\%, \textit{while keeping other features unchanged}, the email will be classified as ``Spam" rather than ``Ham". Hence, the less the number of features need to change from $\pmb{x}$ to generate $\mathbf{\tilde{x}}$, the more ``concise" the explanation becomes. To achieve this goal, we add the constraint as 
\begin{equation}
        \left|\mathcal{S}\right| \leq \textit{K}
        \label{cond2}
\end{equation}
where $\mathcal{S}$ is the feature set of $\pmb{x}$ that are perturbed to generate $\tilde{\pmb{x}}$, hence making $\left|\mathcal{S}\right|$ the number of features changed, i.e.,  
\begin{equation}
    \left|\mathcal{S}\right| = \sum_{m=1}^M\mathbbm{1}(\mathbf{x^m} \not= \mathbf{\tilde{x}^m})
\end{equation}

We not only want to change a minimum number of features, but also want those features to be \textit{informative}. For example, the explanation ``Had the frequency of `!' and `!!!' is more than 0.3, the email would be classified as spam rather than ham" is not as informative as ``Had the frequency of `!' and `wonder' is more than 0.3, the email would be classified as spam rather than ham." Hence, we want $\mathcal{S}$ to contain a list of perturbed features such that any pairwise mutual information among them is within an upper-bound $\gamma$. Thus, we add the constraint:
\begin{equation}
    \text{SU}(\mathcal{X}^{\mathit{i}}, \mathcal{X}^{\mathit{j}})  \leq \gamma \quad \bigforall \mathit{i},\mathit{j} \in \mathcal{S}
    \label{cond3}
\end{equation}
where $\mathrm{SU}(\cdot)$ is \textit{Symmetrical Uncertainty} function, a normalized form of mutual information, to be introduced in Section \ref{sec:feature_selection}. 
Finally, we also need to ensure that the final predicate $\mathcal{P}$ is realistic. For example, the age feature should be a positive integer). Therefore, we want to generate $\mathbf{\tilde{x}}$ such that it conforms to features domain constraints of the dataset:
\begin{equation}
    \mathbf{\tilde{x}} \in dom(\mathcal{X})
    \label{cond4}
\end{equation}

Newly introduced constraints are novel from previous adversarial literature, which focus more on minimizing the difference $\Vert\pmb{x} - \mathbf{\tilde{x}}\Vert_p$. However, minimizing such a distance alone will not necessarily make $\mathbf{\tilde{x}}$ more self-explanatory to users. Instead, we propose that as long as the constraints on the maximum number of perturbed features, i.e., Eq. (\ref{cond2}), their entropy, i.e., Eq. (\ref{cond3}), and domain, i.e., Eq. (\ref{cond4}), is satisfied, we can generate more concise and informative explainable contrastive samples. From the above analysis, we formalize the objective function as follows.

\noindent
\vspace{-25pt}
\begin{center}
\fbox{\parbox[t]{0.9\linewidth}{
\textbf{\textsc{Objective Function}}: Given $\pmb{x}$, hyperparameter $\mathit{K}$, $\gamma$, our goal is to generate new contrastive sample $\mathbf{\tilde{x}}$ to explain the prediction $\mathit{f}(\pmb{x})$ by solving the objective function:
\begin{equation}
\begin{aligned}
        \min_{\tilde{\pmb{x}}} & \quad dist(\tilde{\pmb{x}}, \pmb{x})\\ 
        s.t.& \quad \mathrm{argmax}(f(\pmb{x})) \ne \mathrm{argmax}(f(\tilde{\pmb{x}})), \quad 
        \left|\mathcal{S}\right| \leq \textit{K}\\
        &  \quad \text{SU}(\mathcal{X}^{\mathit{i}}, \mathcal{X}^{\mathit{j}}) \leq \gamma \quad \bigforall \mathit{i},\mathit{j} \in \mathcal{S}, \quad \mathbf{\tilde{x}} \in dom(\mathcal{X})  
\end{aligned}
\end{equation}
\label{objective}
}}
\end{center}
\vspace{-10pt}

\section{\mymethod: GENERATING INTERVENTIVE CONTRASTIVE SAMPLES FOR MODEL EXPLANATION}\label{sec:generation}

\begin{figure}[t!]
  \vspace*{-10pt}
  \centering
  \includegraphics[width=0.4\textwidth]{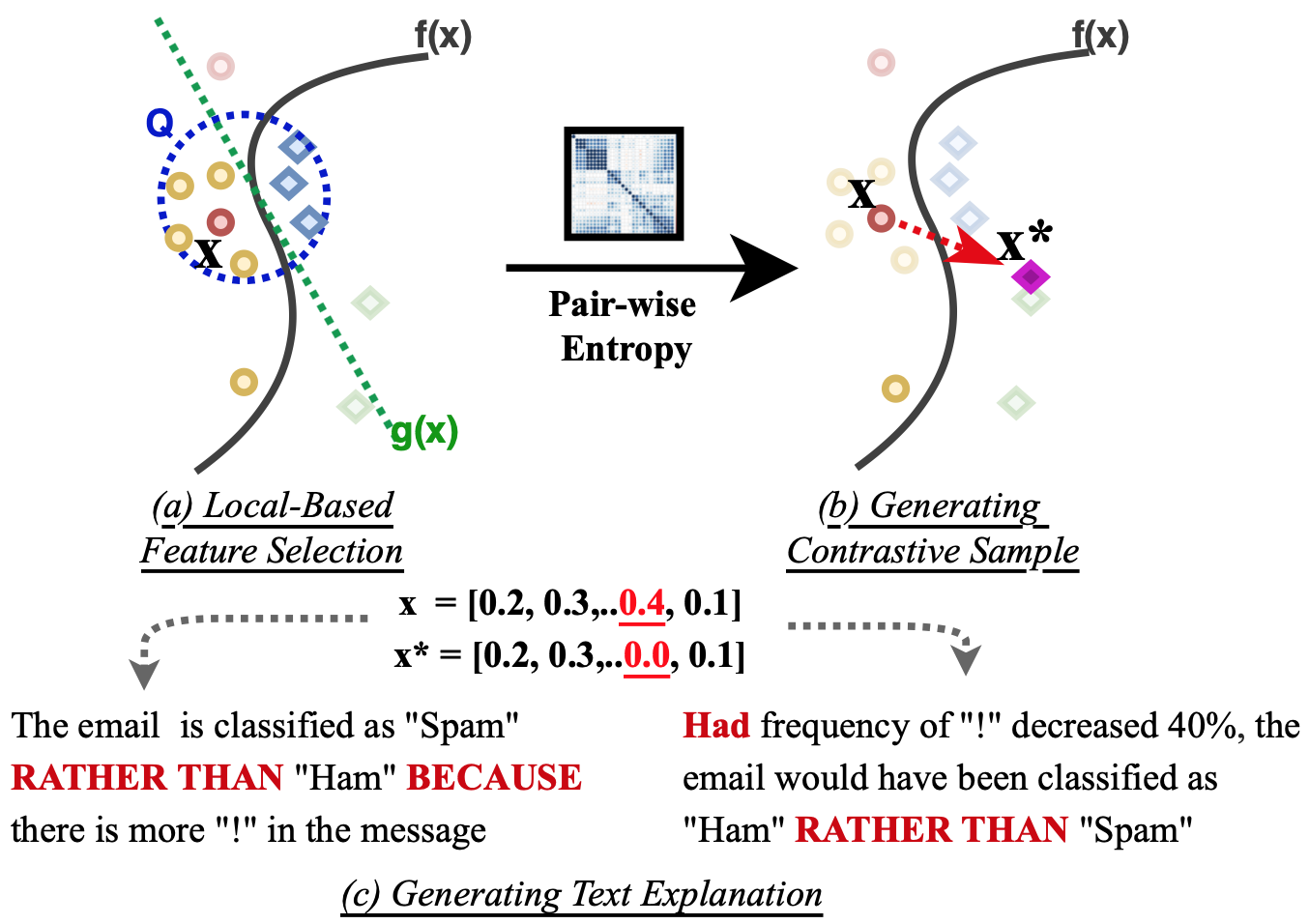}
  \caption{\textsc{\mymethod} with \textsc{Local-Based} Feature Ranking}
  \label{fig:workflow}
  \vspace*{-0.6cm}
\end{figure}

This section describes how to solve the objective function and the details of \textsc{\mymethod}. Figure ~\ref{fig:workflow} gives an illustration of the framework. It consists of three steps: (i) entropy-based forward features ranking, which aims at finding instance-dependent features satisfying the constraint; (ii) generate contrastive samples with the selected features; and (iii) create an explanation text based on generated sample $\tilde{\pmb{x}}$. Alg. \ref{alg:grcs} describes \textsc{\mymethod} algorithm.

\subsection{Contrastive Sample Generation Algorithm}
Before introducing how to obtain a list of potential features to perturb, in this section, we first describe our contrastive sample generation algorithm by assuming that the ordered feature list $\mathcal{U}^*$ is given. To solve $\tilde{\pmb{x}}$ such that Eq. (\ref{cond1}) is satisfied, we can continuously perturb $\tilde{\pmb{x}}$ by projecting itself on the decision hyperplane separating it with the nearest contrastive class $v$. Particularly, at each time-step i, we project $\tilde{\pmb{x}}$ with orthogonal projection vector $\mathbf{r}_v$:
\begin{equation}
    \mathbf{r}_v= \frac{|\mathit{f}_v(\tilde{\pmb{x}}_{i-1}) - \mathit{f}_C(\pmb{x})|}{\Vert\nabla\mathit{f}_v(\tilde{\pmb{x}}_{i-1}) - \nabla\mathit{f}_C(\pmb{x})\Vert_2^2}\;\Big(\nabla\mathit{f}_v(\tilde{\pmb{x}}_{i-1}) - \nabla\mathit{f}_C(\pmb{x})\Big)
\label{eqn:gen:2}
\end{equation}
\noindent where $f_v(\tilde{\pmb{x}}_{i-1})$ is the confidence of $f$ on $\tilde{\pmb{x}}_{i-1}$ being classified as class $v$. $C \longleftarrow \mathrm{argmax}(\mathit{f}(\pmb{x}))$ is the current prediction label, and contrastive class $\pmb{v}$ can be inferred with Alg. \ref{alg:generation}, Ln. \ref{alg:grcs:contrastive}. Intuitively, $\pmb{v}$ is the contrastive class across the \textit{closest} hyperplane of the decision boundary from $\pmb{x}$.

\begin{algorithm}[t]
    \caption{\textsc{\mymethod}}
    \label{alg:grcs}
    \begin{flushleft}
        \hspace*{\algorithmicindent}\textbf{Input:} $\mathit{f}, \pmb{x}$, $\mathit{K}$, $\gamma$, $\mathcal{X}$\\
        \hspace*{\algorithmicindent}\textbf{Output: $\tilde{\pmb{x}},\;\tilde{\mathbf{y}}$}
    \end{flushleft}
    \begin{algorithmic}[1]
        \STATE \textit{Initialize: } $\tilde{\pmb{x}} \longleftarrow \pmb{x}$, $j \longleftarrow 1$
        \STATE $\mathcal{U} \longleftarrow$ $\mathit{W}_{Gradient}(\mathit{f}, \pmb{x})$  \textbf{OR} $\mathit{W}_{Local}(\mathit{f}, \pmb{x})$ \label{alg:grcs:while0}
        \STATE $\mathcal{U}^* \longleftarrow$ $\mathit{W}_{Entropy}(\mathcal{X},\mathcal{U}, \gamma)$ \label{alg:grcs:while2}
        \WHILE{$\mathrm{argmax}(\mathit{f}(\tilde{\pmb{x}}))=\mathrm{argmax}(\mathit{f}(\pmb{x}))$ \textbf{AND} $k \leq \mathit{K}$} \label{alg:grcs:while}
            
            \STATE $\tilde{\pmb{x}} \longleftarrow$ \textsc{GenerateContrastiveSample}($\mathit{f},\pmb{x},\mathcal{U}^*[:k]$)
            \STATE $k \longleftarrow k+1$
        \ENDWHILE
        \STATE $\tilde{\mathbf{y}} \longleftarrow \mathrm{argmax}(\mathit{f}(\tilde{\pmb{x}}))$
        \STATE \textbf{Return} $\tilde{\pmb{x}},\;\tilde{\mathbf{y}}$
    \end{algorithmic}
\end{algorithm}

\begin{algorithm}[th]
    \caption{\textsc{GenerateContrastiveSample}}
    \label{alg:generation}
    \begin{flushleft}
        \hspace*{\algorithmicindent}\textbf{Input:} $\mathit{f}, \pmb{x}$, $\mathcal{S}$\\
        \hspace*{\algorithmicindent}\textbf{Output: $\tilde{\pmb{x}}$}
    \end{flushleft}
    \begin{algorithmic}[1]
        \STATE \textit{Initialize: }$\mathbf{\tilde{x}_0} \longleftarrow \pmb{x}$, $i\longleftarrow0$, $C \longleftarrow \mathrm{argmax}(\mathit{f}(\pmb{x}))$
        \STATE $v \longleftarrow \mathrm{argmin}_{c\not=C}\frac{|\mathit{f}_c(\pmb{x}) - \mathit{f}_{C}(\pmb{x})|}{||\nabla\mathit{f}_c(\pmb{x}) - \nabla\mathit{f}_{C}(\pmb{x})||_2}$ \label{alg:grcs:contrastive}
        \WHILE{$\mathrm{argmax}(\mathit{f}(\tilde{\pmb{x}}_i))=\mathrm{argmax}(\mathit{f}(\pmb{x})) \;\mathrm{\textbf{AND}}\; i<\mathrm{STEPS} $}\label{alg:generation:start}
            \STATE $\mathbf{r}_v= \frac{|\mathit{f}_v(\tilde{\pmb{x}}_{i-1}) - \mathit{f}_C(\pmb{x})|}{\Vert\nabla\mathit{f}_v(\tilde{\pmb{x}}_{i-1}) - \nabla\mathit{f}_C(\pmb{x})\Vert_2^2}\;(\nabla\mathit{f}_v(\tilde{\pmb{x}}_{i-1}) - \nabla\mathit{f}_C(\pmb{x}))$
            \STATE $\tilde{\pmb{x}}_{i+1}[\mathcal{S}] \longleftarrow \tilde{\pmb{x}}_{i}[\mathcal{S}] + \pmb{r}_v[\mathcal{S}] $
            \STATE $\tilde{\pmb{x}}_{i+1} \longleftarrow$ $\mathit{P}(\tilde{\pmb{x}}_{i+1},\mathit{dom}(\mathcal{X}))$ \label{alg:generation:end} \label{alg:generation:project}
            \STATE $i\longleftarrow i+1$
        \ENDWHILE
        \STATE \textbf{Return} $\tilde{\pmb{x}}$
    \end{algorithmic}
\end{algorithm}

To address constraint Eq. (\ref{cond2}), instead of perturbing all of the features, we update $\tilde{\pmb{x}}$ only on the first $k\leq \mathit{K}$ features from an ordered list $\mathcal{U}^*$, which will be introduced later, at each time step $i$ until it crosses the decision boundary:
\begin{equation}
\begin{aligned}
    \mathcal{S} \longleftarrow \mathcal{U}^*[:k], \quad
    \tilde{\pmb{x}}_i[\mathcal{S}] \longleftarrow \tilde{\pmb{x}}_{i-1}[\mathcal{S}] + \mathbf{r}_v[\mathcal{S}]
\end{aligned}
\label{eqn:gen:1}
\end{equation}
Since feature perturbation based on $\mathbf{r}_v$ does not always guarantee that resulted $\tilde{\pmb{x}}_i$ still maintains in the original feature space, to address constraint Eq. (\ref{cond4}), we project those adjusted features back on to the original domain of $\mathcal{X}$:
\begin{equation}
    \tilde{\pmb{x}}_i \longleftarrow \mathit{P}(\tilde{\pmb{x}}_i,\mathit{dom}(\mathcal{X}))
    \label{eqn:gen:3}
\end{equation}
where $P$ is a projection which ensures that final $\mathbf{\tilde{x}}$ looks more real (e.g., age feature should be a whole number and > 0). The domain space $\mathit{dom}(\mathcal{X})$ can include the maximum, minimum, and data types (e.g., int, float, etc.) of each feature. These can be either calculated from the original training set or manually set by domain experts. 

With a fixed $k$, Eq. (\ref{eqn:gen:3}) does not guarantee that $\tilde{\pmb{x}}$ will always cross the decision boundary to class $\pmb{v}$. Hence, we gradually increase $k \longrightarrow \mathit{K}$ until a contrastive sample is successfully generated, i.e., $\mathrm{argmax}(f(\pmb{x})) \ne \mathrm{argmax}(f(\tilde{\pmb{x}}_i))$ or when $k==\mathit{K}$. Alg. \ref{alg:generation} illustrates the steps to generate contrastive samples.

One obvious challenge is how to come up with the ordered list $\mathcal{U}^*$ of features to perturb. Next, we will describe this in detail.

\subsection{Entropy-Based Forward Feature Ranking}\label{sec:feature_selection}

As different $\pmb{x}$ might require a different subset of features to perturb, the first challenge is how to prioritize features that are highly vulnerable to the contrastive class $\pmb{v}$. To do this, we rank all features of $\pmb{x}$ according to their predictive power w.r.t prediction $\mathit{f}(\pmb{x})$, resulting in an ordered list $\mathcal{U}$:
\begin{equation}
    \mathcal{U} \longleftarrow \mathit{W}(\mathit{f}, \pmb{x})
\end{equation}
where $\mathit{W}(\cdot)$ is a feature ranking function. The most straight forward way is to rank all features according to their gradients w.r.t the \textit{nearest} contrastive class $\pmb{v}$ that back-propagates through $\mathit{f}(\pmb{x})$, resulting in $\mathit{W}_{\textsc{Gradient}}(\mathit{f}, \pmb{x})$ that returns the ranking of the following set:
\begin{equation}
    \{\nabla\mathit{f}_{\pmb{v}}(\pmb{x}^1), \nabla\mathit{f}_{\pmb{v}}(\pmb{x}^2),..\nabla\mathit{f}_{\pmb{v}}(\pmb{x}^M)\}
\end{equation}
While this method is straightforward, these gradients capture a global view of feature rankings, rather than being customized to a local vicinity of decision boundary around $\pmb{x}$. To overcome this limitation, we introduce $ \mathit{W}_{\textsc{Local}}(\mathit{f}, \pmb{x})$ to return the ranking of the following set:
\begin{equation}
    \{w^1_{\mathit{g}(\pmb{x})}, w^2_{\mathit{g}(\pmb{x})},...w^M_{\mathit{g}(\pmb{x})}\}
\end{equation}
with $w^j_{\mathit{g}(\pmb{x})}$ is the feature importance score of the $j$-th feature  returned from an interpretable ML model $\mathit{g}(\pmb{x})$ (e.g., feature weights for logistic regression, Gini-score for decision tree, etc.). $\mathit{g}(\cdot)$ is trained on a subset of data points $\mathcal{Q}$ surrounding $\pmb{x}$ (Figure \ref{fig:workflow}a) with \textit{maximum likelihood estimation (MLE)} as the loss function:
\begin{equation}
    \min_{\theta_{\mathit{g}(x)}} \frac{1}{|\mathcal{Q}|}\sum_{\pmb{x}\in\mathcal{Q}}\mathit{f}(\pmb{x})\;\mathrm{log}(\mathit{g}(\pmb{x}))
\end{equation}
If the prediction of $\mathit{g}(\pmb{x})$ on $\mathcal{Q}$ is very close to that of $\mathit{f}(\pmb{x})$, important features from $\mathit{g}(\pmb{x})$ are more prone to change in $\mathit{f}(\pmb{x})$. $\mathcal{Q}$ can be collected by sampling $q$ nearest data points to $\pmb{x}$ from \textit{each of the predicted classes} by $\mathit{f}(\pmb{x})$ on the training set using \textit{NearestNeighbors(NN)} search algorithm with different distance functions. We set $q=4$ and use \textit{Euclidean distance} throughout all experiments. 

In the aforementioned variants of $\mathit{W}(\cdot)$, each feature is treated independently with each other. However, if a pair of selected features are highly dependent on each other (e.g., frequency of "!" and "!!"), the final generated samples will be less informative. Because of this, also to address constraint Eq.(\ref{cond3}), we want to generate a new ordered list $\mathcal{U}^*$ as follows:
\begin{equation}
    \mathcal{U}^* \longleftarrow \mathit{W}_{Entropy}(\mathcal{X}, \mathcal{U})
\end{equation}
where $\mathit{W}_{Entropy}$ is a forward-based selection approach, which will \textit{iteratively} add each feature from $\mathcal{U}$ (from the most to least predictive) to $\mathcal{U}^*$ one by one such that the mutual information of \textit{any} pairs of features in $\mathcal{U}^*$ is within a upper-bound $\gamma$:
\begin{equation}
      SU(\mathcal{X}^{\mathit{i}}\,,\,\mathcal{X}^{\mathit{j}}) \leq \gamma \quad \bigforall \mathit{i},\mathit{j} \in\mathcal{U}^*
\end{equation}
where $SU \in [0, 1]$ is an entropy-based \textit{Symmetrical Uncertainty} \cite{flannery1988numerical} function that measures the mutual information between the $i$-th and $j$-th feature of $\mathcal{X}$ as:
\begin{equation}
    SU(\mathcal{X}^\mathit{i}\,,\,\mathcal{X}^\mathit{j}) = 2\left[\frac{IG(\mathcal{X}^\mathit{i}\,|\,\mathcal{X}^\mathit{j})}{H(\mathcal{X}^\mathit{i}) + H(\mathcal{X}^\mathit{j})}\right]
\end{equation}

\noindent where $IG(\mathcal{X}^\mathit{i}\,|\,\mathcal{X}^\mathit{j})$ is the \textit{information gain} of $\mathcal{X}^\mathit{i}$ given $\mathcal{X}^\mathit{j}$, and $H(\mathcal{X}^\mathit{i})$, $H(\mathcal{X}^\mathit{j})$ is empirical entropy of $\mathcal{X}^\mathit{i}$ and $\mathcal{X}^\mathit{j}$, respectively. A value $SU = 1$ indicates that $\mathcal{X}^\mathit{i}$ completely predicts the value of $\mathcal{X}^\mathit{j}$ \cite{yu2003feature}. In implementation, we first \textit{normalize} $\mathcal{X}$ and adapt multi-interval discretization \cite{fayyad1993multi} to convert continuous features to discrete format before applying $SU(\cdot)$. Alg. \ref{alg:grcs:entropy} describes function $\mathit{W}_{Entropy}$ in details.
One obvious challenge is how to come up with the ordered list $\mathcal{U}^*$ of features to perturb. The next section will describe this in detail.

\vspace*{-5pt}
\begin{algorithm}[th]
    \caption{Feature Selection with Entropy: $\mathit{W}_{Entropy}$}
    \label{alg:grcs:entropy}
    \begin{flushleft}
        \hspace*{\algorithmicindent}\textbf{Input:} $\mathcal{X}, \mathcal{U}$, $\gamma$\\
        \hspace*{\algorithmicindent}\textbf{Output:} Ordered list of features $\mathcal{U}^*$\\
    \end{flushleft}
    \begin{algorithmic}[1]
        \STATE \textit{Initialize: } $\mathcal{U}^* \longleftarrow \{\}$, $\hat{\mathcal{X}} \longleftarrow \mathrm{normalize}(\mathcal{X})$
        \FOR{$i$ \textbf{in} $\mathcal{U}$}
            \STATE $to\_add \longleftarrow True$
                \FOR{$j$ \textbf{in} $\mathcal{U}^*$}
                    \IF{\textsc{SU}($\hat{\mathcal{X}}^i, \hat{\mathcal{X}}^j$) $> $ $\gamma$} $to\_add \longleftarrow False$
                    \ENDIF
                \ENDFOR
            \IF{$to\_add = True$}  $\mathcal{U}^* \longleftarrow \mathcal{U}^* \cup \{i\}$
            \ENDIF
        \ENDFOR
        \STATE \textbf{Return} $\mathcal{U}^*$
    \end{algorithmic}
\end{algorithm}
\vspace*{-2pt}

\begin{table*}[th!]
\caption{Examples of generated contrastive samples and their explanation texts}
\label{tab:explain}
\vskip -1em
\small
\begin{tabular}{l|llrrl|l}
\toprule
Dataset &  Features/\textit{Prediction}  & Type & Original & Generated & Changes & Explanation Text \\
\hline
\hline
\multirow{2}{*}{cancer95} & bare\_nuclei & int & 1 & 10 & $\boldsymbol{\uparrow}$ 9 & \multirow{2}{8cm}{"if there were 9 more bare nucleus, the patient would be classified as malignant \textit{RATHER THAN} benign"}\\
{} & \textit{diagnosis} & & benign & malignant &\\
\hline
\multirow{3}{*}{spam} & word\_freq\_credit & float & 0.470 & 0.225 & $\boldsymbol{\downarrow}$ 0.245 & \multirow{3}{8cm}{"The message is classified as spam \textit{RATHER THAN} ham because the word 'credit' and 'money' is used twice as frequent as that of ham message"}\\
{} & word\_freq\_money & float & 0.470 & 0.190 & $\boldsymbol{\downarrow}$ 0.280 \\
{} & \textit{class} & & Spam & Ham &\\
\bottomrule
\end{tabular}
 \vskip -1em
\end{table*}

\vspace*{-15pt}
\subsection{Generating Explanation Text}

After generating contrastive sample $\tilde{\pmb{x}}$, we take a further step and generate an explanation in natural text. Table \ref{tab:explain} shows generated contrastive samples and the corresponding explanation for various datasets with \textit{$\mathit{K}$ = 5}. To do this, for a specific prediction $\mathit{f}(\textbf{x})$ and generated contrastive sample $\tilde{\textbf{x}}$, we first calculate their feature differences, resulting in predicate $\mathcal{P}$ as defined in Def. \ref{def:intervention}. Then, we can translate $\mathcal{P}$ to text by using condition-based text templates such as {\ttfamily ... is classified as X RATHER THAN Y because ...}, or {\ttfamily had..., it would have been classified as X RATHER THAN Y} (Figure \ref{fig:workflow}c). Different text templates can be selected randomly to induce diversity in explanation text. The difference in features values can be described in three different degrees of obscurity from (i) \textit{extract value} (e.g., {\ttfamily 0.007 point lower}), to (ii)  \textit{magnitude comparison} (e.g., {\ttfamily twice as frequent}), or (iii) \textit{relative comparison} (e.g., {\ttfamily higher, lower}). Which degree of detail to best use is highly dependent on the specific feature, domain, and the choice of end-users, and they do not need to be consistent among perturbed features in a single explanation text.




\subsection{Complexity Analysis}
According to Alg. \ref{alg:grcs}, we analyze the time complexity of {\sc \mymethod} on each prediction instance as follows. The predictive feature ranking step using $\mathcal{W}_{Gradient}$ takes $\mathcal{O}(M \log M)$ with \textit{Quick Sort}. Reordering the ranked list of features with $\mathit{W}_{Entropy}$ takes $\mathcal{O}(M)$. Generating contrastive sample step takes $\mathcal{O}(M) + Z\mathcal{V}$ with $K \ll M$, where $Z$ is total number of classes to predict, and $\mathcal{V}$ is the time complexity of forward and backward pass of $\mathit{f}(\pmb{x})$. Generating an explanation text then takes another $\mathcal{O}(M)$. To sum up, the overall time complexity of {\sc \mymethod} to generate an explanation for each prediction instance is $\mathcal{O}(M \log M) + Z\mathcal{V}$ with $K \ll M$ (excluding the overhead of training $\mathit{g}(\cdot)$ and searching for $\mathcal{Q}$ in case of $\mathit{W}_{Local}$).

\section{EXPERIMENTS}\label{sec:experiment}
In this section, we conduct experiments to verify the effectiveness of \textsc{\mymethod}. Specifically, we want to answer two questions: (i) how good are the generated interventive contrastive samples? and (ii) how good are the generated explanation?

\begin{table}[t!]
\small
\caption{Dataset statistics and prediction performance}
\label{tab:dataset}
\vskip -1em
\begin{tabular}{l|lrrrHHH|cc}
    \toprule
    Size & Dataset & \#Class & \#Feat. & \#Data & \# of Train & \# of Validation & \# of Test & $Acc.^*$ & $F1^*$ \\
    \hline
    \multirow{6}{*}{small} & eegeye                       & 2             & 14             & 14980         & 12133         & 1349               & 1498         & 0.858            & 0.858       \\
    & diabetes                     & 2             & 8              & 768           & 621           & 70                 & 77           & 0.779             & 0.777       \\
    & cancer95                     & 2             & 9              & 699           & 566           & 63                 & 70           & 0.963             & 0.963       \\
    & phoneme                      & 2             & 5              & 5404          & 4376          & 487                & 541          & 0.774             & 0.772       \\
    & segment                      & 7             & 19             & 2310          & 1871          & 208                & 231          & 0.836             & 0.817       \\
    & magic                        & 2             & 10             & 19020         & 15406         & 1712               & 1902         & 0.862             & 0.859       \\
    \hline
    \multirow{3}{*}{medium} & biodeg         &       2 &      41 &   1055  &  854   &   95   &    106  & 0.853 & 0.851\\
    & spam                         & 2             & 57             & 4601          & 3726          & 414                & 461          & 0.932             & 0.932       \\
    & cancer92                       & 2             & 30             & 569           & 460           & 52                 & 57           & 0.958             & 0.958       \\
    \hline
    \multirow{2}{*}{large} & mfeat                        & 10            & 216            & 2000          & 1620          & 180                & 200          & 0.943             & 0.936      \\
    & musk                         & 2             & 166            & 476           & 385           & 43                 & 48           & 0.783             & 0.789       \\
    \bottomrule
    \multicolumn{10}{l}{(*) Accuracy and F1 scores are averaged across 10 different runs.}
    \end{tabular}
    \vskip -0.7cm
\end{table}

\subsection{Datasets}
We select 11 publicly available datasets of different domains and scales from \cite{uci} to fully evaluate and understand how well {\sc \mymethod} works with neural networks trained on data with varied properties. As shown in Table.\ref{tab:dataset}, the datasets are grouped into three scale levels according to the number of features. Each dataset is split into training, validation, and test set with a ratio of 8:1:1, respectively. The table also includes the performance of different neural models (Appendix \ref{sec:neural_models}) on test set in both Accuracy and F1 score.


\subsection{Compared Methods}
Since our proposed framework combines the best of both worlds: adversarial generation and neural network model explanation, we select various relevant baselines from two aspects.
\begin{itemize}[leftmargin=*]
    \item {\sc NearestCT}: Instead of generating synthetic contrastive sample for explanation for data point $\pmb{x}$, this approach selects the \textit{nearest} contrastive samples of $\pmb{x}$ from the \textit{training set} to provide contrastive explanation for the prediction $f(\pmb{x})$.
    \item {\sc DeepFool}\cite{moosavi2016deepfool}: An effective approach that was originally proposed for untargeted attack by generating adversarial samples. Even though {\sc DeepFool} is not designed for generating samples to explain predictions, we consider this as a baseline that intervenes on \textit{all features} to generate contrastive samples.
    \item {\sc Lime} \cite{lime}: A \textit{local interpretable model-agnostic explanation} approach that provides explanation for individual prediction. This approach replies on visualization of feature importance scores (for text and tabular data), and feature heat-map (for image data) to deliver explanation. We use an out-of-the-box toolkit\footnote{\url{https://github.com/marcotcr/lime}} to run experiments for comparison. {\sc Lime} is selected mainly due to its popularity as a baseline for ML explanation approach.
    \item \textsc{\mymethod-Gradient} and \textsc{\mymethod-Local} (\textbf{ours}): \textsc{\mymethod} with predictive feature ranking function $\mathit{W}$ as $\mathit{W}_{Gradient}$ and $\mathit{W}_{Local}$, respectively.
\end{itemize}

\begin{table*}[th!]
\centering
\small
\caption{All results are averaged across 10 different runs. The best and second best results are highlighted  in \textbf{bold} and \underline{underline}.}
\vskip -1em
\label{tab:result}
\begin{tabular}{@{}c|c|rrrrrrrrrrr@{}}
\toprule
\multirow{2}{*}{Statistics} & \multirow{2}{*}{Dataset} & \multicolumn{6}{c}{\# Features < 30} & \multicolumn{3}{c}{30 $\leq$ \# Features < 100} & \multicolumn{2}{c}{100 $\leq$ \# Features}\\
\cmidrule(l{5pt}r{5pt}){3-8}  \cmidrule(l{5pt}r{5pt}){9-11} \cmidrule(l{5pt}r{5pt}){12-13}
{} & {} & \textbf{eegeye} & \textbf{diabetes}  & \textbf{cancer95} & \textbf{phoneme} & \textbf{segment} & \textbf{magic} & \textbf{biodeg} & \textbf{spam} & \textbf{cancer92} & \textbf{mfeat} & \textbf{musk} \\ 
\hline

\multirow{4}{*}{$\mathbf{R}_{\mathrm{avg\#Feats}}$} & {\sc NearestCT}      &             13.56 &              6.93 &             5.92 &             4.82 &             16.10 &              9.97 &             20.53 &             17.50 &            29.97 &            204.22 &            147.86 \\
& {\sc DeepFool}       &             14.00 &              8.00 &             9.00 &             5.00 &             19.00 &             10.00 &             41.00 &             57.00 &            30.00 &            216.00 &            166.00 \\
& \textsc{\mymethod-Local}     &  \underline{1.15} &     \textbf{1.55} &  \underline{2.7} &    \textbf{1.25} &     \textbf{2.42} &  \underline{1.68} &  \underline{3.07} &  \underline{2.95} &    \textbf{3.95} &  \underline{3.28} &  \underline{3.74} \\
& \textsc{\mymethod-Gradient} &      \textbf{1.0} &  \underline{1.96} &    \textbf{2.66} &  \underline{1.3} &  \underline{3.84} &      \textbf{1.6} &     \textbf{1.93} &     \textbf{1.09} &  \underline{4.5} &     \textbf{2.76} &     \textbf{2.85} \\

\hline
\multirow{4}{*}{$\mathbf{R}_{\mathrm{info-gain}}^*$} & {\sc NearestCT}      &  \underline{0.69} &              0.44 &     \textbf{0.64} &              0.12 &              0.19 &              0.04 &              0.44 &  \underline{0.62} &              0.02 &  \underline{0.58} &             0.28 \\
& {\sc DeepFool}       &      \textbf{0.7} &              0.41 &  \underline{0.62} &              0.12 &  \underline{0.33} &              0.05 &  \underline{0.58} &              0.53 &              0.01 &     \textbf{0.59} &             0.29 \\
& {\sc \mymethod-Local}     &             0.64 &     \textbf{0.79} &              0.49 &     \textbf{0.81} &     \textbf{0.55} &  \underline{0.67} &              0.46 &              0.47 &     \textbf{0.13} &              0.34 &  \underline{0.3} \\
& {\sc \mymethod-Gradient} &              0.64 &  \underline{0.62} &              0.52 &  \underline{0.78} &              0.23 &     \textbf{0.71} &     \textbf{0.76} &     \textbf{0.95} &  \underline{0.04} &              0.50 &     \textbf{0.4} \\

\hline

\multirow{4}{*}{$\mathbf{R}_{\mathrm{influence}}$} & {\sc NearestCT}      &              0.05 &              0.06 &              0.11 &              0.03 &              0.01 &             0.00 &              0.02 &              0.04 &              0.00 &             0.00 &              0.00 \\
& {\sc DeepFool}       &              0.05 &              0.05 &              0.07 &              0.02 &              0.02 &             0.00 &              0.01 &              0.01 &              0.00 &             0.00 &              0.00 \\
& {\sc \mymethod-Local}     &  \underline{0.55} &     \textbf{0.52} &  \underline{0.18} &     \textbf{0.65} &     \textbf{0.23} &  \underline{0.4} &  \underline{0.15} &  \underline{0.16} &     \textbf{0.04} &  \underline{0.1} &  \underline{0.08} \\
& {\sc \mymethod-Gradient} &     \textbf{0.64} &  \underline{0.33} &      \textbf{0.2} &  \underline{0.61} &  \underline{0.06} &    \textbf{0.45} &      \textbf{0.4} &     \textbf{0.88} &  \underline{0.01} &    \textbf{0.18} &     \textbf{0.14} \\

\bottomrule
\end{tabular}
\vskip -1.2em
\end{table*}
\vspace{-0.1cm}
\subsection{Evaluation of Generated Samples}\label{sec:quantify}
In this section, we want to examine the quality of generated contrastive samples. Since {\sc DeepFool}, {\sc NearestCT} and {\sc \mymethod} generate intermediate samples to explain predictions, while {\sc Lime} is not, we compare and analyze {\sc Lime} separately in Section \ref{sec:case_study} to evaluate final generated explanation.

For each dataset, we train a neural network model $\mathit{f}(\cdot)$ using the training set. We tune it using the validation set together with early-stopping strategy to prevent overfitting and report its performance on the test set. Table~\ref{tab:dataset} reports the averaged performance across 10 different runs. We set $\mathit{K}=5,\mathit{\gamma}=0.5$, and generate $\tilde{\pmb{x}}$ to explain predictions $\mathit{f}(\pmb{x})$ of all samples in test set, resulting in the set of generated contrastive samples $\tilde{\mathcal{X}}$.

To thoroughly examine the proposed approach, we come up with the following analytical questions (AQs).

\begin{enumerate}[label=\textbf{AQ\arabic*},noitemsep,align=left,leftmargin=*]
    \item \textbf{Fidelity}: How accurate are the generated contrastive samples' labels, i.e., whether they can cross neural network model's decision boundary as expected? \label{aq:fidelity}
    \item \textbf{Conciseness}: How concise are generated samples, i.e., how many features needed to be perturbed to successfully generate contrastive samples? \label{aq:concise}
    \item \textbf{Info-gain}: How informative are generated samples? \label{aq:reasonableness} 
    \item \textbf{Influence}: Derived from Def. \ref{def:influencescore}, how well do the generated samples answer the question \textit{Why X rather than Y?}.
    \label{aq:explainability}
\end{enumerate}

\subsubsection{\ref{aq:fidelity} (Fidelity)}

Fidelity, measured by $\mathbf{R}_{\mathrm{fidelity}}$, shows how accurately contrastive samples are generated according to the neural network model's boundary, i.e., the accuracy of generated samples' labels w.r.t their predictions by the neural network model:
\begin{equation}
    \mathbf{R}_{\mathrm{fidelity}} = \frac{1}{|\tilde{\mathcal{X}}|}\sum_{(\mathbf{\tilde{x}}, \mathbf{\tilde{y}}) \in \tilde{\mathcal{X}}}\mathbbm{1}(\mathbf{\tilde{y}} = \mathrm{argmax}(\mathit{f}(\mathbf{\tilde{x}})))
\end{equation}

\begin{figure}[h!]
  \centering
  \hspace{-5pt}
  \vskip -1em
  \includegraphics[width=0.45\textwidth]{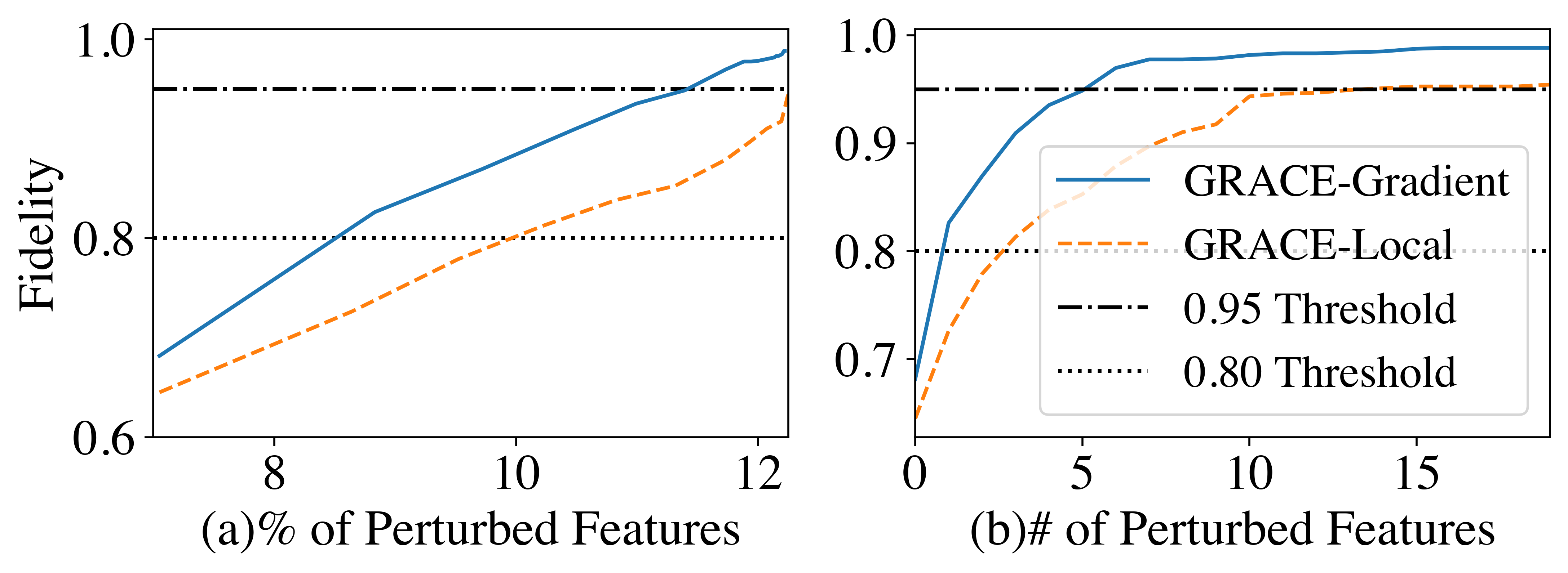}
   \vskip -1em
  \caption{Percentage of perturbed features v.s fidelity}
  \label{fig:concise}
 \vskip -1em
\end{figure}

Different from the two baselines, two variants of \textsc{\mymethod} have to satisfy the domain constraints, minimize the number of features, and their entropy, all at the same time. 
Nevertheless, with $K=5$, our method shows an average $\mathbf{R}_{\mathrm{fidelity}}$ of around 0.8 for both \textsc{\mymethod-Gradient} and \textsc{\mymethod-Local}. As the \# of perturbed features increases, the Fidelity scores for both \textsc{\mymethod-Gradient} and \textsc{\mymethod-Local} also increase, which satisfies the expectation (Figure \ref{fig:concise}). 

\subsubsection{\ref{aq:concise} (Conciseness)}
  
We not only want to generate samples with high \textit{fidelity}, but also want to perturb as few features as possible. Thus, we introduce \textit{conciseness} that measures the ability to generate $\tilde{\pmb{x}}$ by changing as few features as possible. To do this, we want to see how \textit{fidelity} correlates with the average number of perturbed features, denoted as $\mathbf{R}_{\mathrm{avg\#Feats}}$:
\begin{equation}
    \mathbf{R}_{\mathrm{avg\#Feats}} = \frac{1}{|\tilde{\mathcal{X}}|}\;\sum_{\mathbf{\tilde{x}} \in \tilde{\mathcal{X}}}|\mathcal{S}_{\tilde{\pmb{x}}}| 
    \label{equa:set_feats}
\end{equation}
where $\mathcal{S}_{\tilde{\pmb{x}}}$ returns the list of features to perturb in $\pmb{x}$ to generate $\tilde{\pmb{x}}$.

Table \ref{tab:result} shows that our {\sc \mymethod} framework dominates the two baselines {\sc DeepFool} and {\sc NearestCT} on $\mathbf{R}_{\mathrm{avg\#Feats}}$ by large margin. Specifically, with $K=5$, our approach is able to generate contrastive samples with much less number of perturbed features, averaging around less than 2.5 features across all datasets. Interestingly, \textsc{\mymethod-Gradient} was able to change on average less than 3 out of a total of 216 and 166 features in the case of \textit{mfeat} and \textit{musk} dataset, respectively.

Moreover, while our method only needs to use as few as 12.5\% of total \# of features to achieve fidelity of around 0.95 (Figure \ref{fig:concise}), {\sc DeepFool} and {\sc NearestCT} baseline needs to change almost 100\% of the total \# of features to achieve a similar score.

\subsubsection{\ref{aq:reasonableness} (Info-gain)}
Since we want to generate samples that are informative, we hope to minimize the averaged mutual information of all pairs of selected features across all samples in $\tilde{\mathcal{X}}$. Hence, we formulate $\mathbf{R}_{\mathrm{\mathrm{info-gain}}}$ to measure such characteristic of being informative as follows:
\begin{equation}
    \mathbf{R}_{\mathrm{info-gain}}  = 1 - \frac{1}{|\tilde{\mathcal{X}}|}\;\sum_{\mathbf{\tilde{x}} \in \tilde{\mathcal{X}}}\;\sum_{\mathit{i} \in \mathcal{S}_{\mathbf{\tilde{x}}}}\sum_{\mathit{j} \in \mathcal{S}_{\mathbf{\tilde{x}}}} \frac{\text{SU}\;(\mathcal{X}^\mathit{i}, \mathcal{X}^\mathit{j})}{|\mathcal{S}_{\mathbf{\tilde{x}}}|^2}
    \label{equa:informative}
\end{equation}

To be fair with other baselines, we instead report $\mathbf{R}_{\mathrm{info-gain}}^*=\mathbf{R}_{\mathrm{info-gain}} \times \mathbf{R}_{\mathrm{fidelity}}$ to take into account the fidelity score.
Even so, thanks to the entropy-aware feature selection mechanism, \textsc{\mymethod} is able to generate contrastive samples that are much more informative compared to {\sc DeepFool}'s in most of the datasets (Table \ref{tab:result}). This shows that samples generated by our framework are not only contrastive but also informative to the final explanation.

\subsubsection{\ref{aq:explainability} (Influence)}
Extended from influence score with tolerance parameter $\lambda=1$ (Def. \ref{def:influencescore}), we aim to measure how well the generated samples can influence the explanation of a specific prediction. Denoted by $\mathbf{R}_{\mathrm{influence}}$, the influence score first captures whether generated samples are still within the original domain space, or $\mathbf{R}_{\mathrm{domain}}$ as follows:
\begin{equation}
        \mathbf{R}_{\mathrm{domain}} = \frac{1}{|\tilde{\mathcal{X}}|}\sum_{\pmb{\tilde{x}} \in \tilde{\mathcal{X}}}\mathbbm{1}(\pmb{\tilde{x}} \in \mathrm{dom}(\mathcal{X}))
        \label{equa:set_domain}
\end{equation}
Moreover, the influence score is also proportional to how faithful generated samples are to the neural network model's decision boundary (\textit{$\mathbf{R}_{\mathrm{fidelity}}$}), how informative they are
($\mathbf{R}_{\mathrm{info-gain}}$), how concise in terms of number of perturbed features ($\mathbf{R}_{\mathrm{avg\#Feats}}$), resulting in a $\mathbf{R}_{\mathrm{influence}}$ calculated as follows:
\begin{equation}
    \mathbf{R}_{\mathrm{influence}} = \frac{\mathbf{R}_{\mathrm{fidelity}} \times \mathbf{R}_{\mathrm{info-gain}} \times \mathbf{R}_{\mathrm{domain}}}{\mathbf{R}_{\mathrm{avg\#Feats}}}
\end{equation}
Intuitively, $\mathbf{R}_{\mathrm{influence}}$ describes the capability to generate new contrastive samples that are both informative, concise, and valid within the original domain space. Hence, the larger the score, the better.

Regarding $\mathbf{R}_{\mathrm{domain}}$, table \ref{tab:result} shows that {\sc DeepFool} performs worst on $\mathbf{R}_{\mathrm{domain}}$, averaged about 0.86, since the generation might move $\mathbf{\tilde{x}}$ much further away from the original distribution, while other methods always ensure that generated samples are within the original domain space. As regards as $\mathbf{R}_{\mathrm{influence}}$, \textsc{\mymethod} is able to generate highly more influential contrastive samples than {\sc DeepFool} and {\sc NearestCT} even when taking $\mathbf{R}_{\mathrm{fidelity}}$ into account, which is the strongest point of both two baselines. 

\captionsetup[figure]{font=small,skip=0pt}

\subsection{Evaluation of Generated Explanation}\label{sec:case_study}
In this section, we want to compare {\sc \mymethod} with {\sc Lime} \cite{lime} from end-users' perspectives on their generated explanation. Before introducing user-studies to compare between two methods, we first draw some observations in a case-study below.

\subsubsection{Case-study: breast-cancer diagnosis}:

 We select \textit{cancer95} dataset to experiment. Following the same experimental setting in Section \ref{sec:experiment}, we apply {\sc Lime} and {\sc \mymethod} on the trained neural network model to explain its predictions on the test set. Figure \ref{fig:lime} depicts explanation produced by {\sc Lime} on a patient diagnosed as malignant by the model. Following guideline published by \textsc{Lime}'s author \footnote{\url{https://github.com/marcotcr/lime}}, explanation for each feature can be interpreted as follows: {\ttfamily "if bare\_nuclei is less than or equal 6.0, on average, this prediction would be 0.15 less malignant"}. With the same prediction, \textsc{\mymethod} generates an explanation text as follows:{\ttfamily "Had bare\_nuclei been 7.0 point lower and clump\_thickness been 9.0 point lower, the patient would have been diagnosed as \textbf{benign rather than malignant}"}

Method wise, both are \textit{instance-based} explanation algorithms, or both explain individual predictions. From Figure \ref{fig:lime}, by presenting top-k important features, {\sc Lime} does not convince if and how a single or combinations of features are vulnerable to the contrastive class, but this is very vivid and concise in case of {\sc \mymethod}. Moreover, while both methods provide some intuition on decisive thresholds at which the prediction would change its direction, the thresholds provided by {\sc Lime} is \textit{only} a local approximation, while that provided by {\sc \mymethod} (e.g., 7.0 point lower for bare\_nuclei) is faithful to the neural network model (\textit{fidelity score} is 1.0). Overall, the explanation generated by {\sc \mymethod} is more concise and faithful to the decision boundary of the neural network model.

\begin{table}[t]
    \small
    \centering
    \caption{User-study with hypothesis testing to compare explanation generated by \textsc{\mymethod} against \textsc{Lime}}
    \vskip -1em
    \begin{tabular}{llccc}
    \hline
    {} & Alternative Hypothesis  &  t-stats &  p-value & df\\
    \hline
    $\mathcal{H}_1$ & \textsc{\mymethod} is more intuitive and friendly & 2.3115 & 0.0104* & 42\\
    $\mathcal{H}_2$ & \textsc{\mymethod} is more comprehensible &  3.0176 & 0.0013** & 42\\
    $\mathcal{H}_3$ & \textsc{\mymethod} leads to more accurate actions & 4.4875 & 3.39$\mathrm{e^{-5}}$** & 37\\
    \bottomrule
    \multicolumn{4}{l}{\textit{*reject Null hypothesis with p-value<0.05 (95\% CI) on one-tailed
    t-test}}\\
    \multicolumn{4}{l}{\textit{**reject Null hypothesis with p-value<0.0 1 (99\% CI) on one-tailed
    t-test}}
    \end{tabular}
    \label{tab:userstudy}
    \vskip -2em
\end{table}

\subsubsection{User-Study 1: Intuitiveness, friendliness \& comprehensibility}: \label{sec:intuitiveness}
We have recruited participants on Amazon Mechanical Turk (AMT) and asked them to compare two explanation methods: \textsc{Lime} and \textsc{\mymethod}. Without assuming or requiring any ML background on the participants, we want to test two alternative hypothesises: explanation generated by {\sc \mymethod} is ($\mathbfcal{H}_1$) more intuitive and friendly, and ($\mathbfcal{H}_2$) more comprehensible than that generated by {\sc Lime} to general users. To test these, using the same prediction instance, we first generate explanation text by \textsc{Lime} and \textsc{\mymethod}. While by default \textsc{Lime} returned explanation for top 10 features, we limit only 5 features that are the most significant. On the contrary, \textsc{\mymethod} only needs 2 features for generating contrastive explanation. Since \textsc{Lime} method originally does not generate explanation text, we then translate its result to text interpretation as described in previous case-study. Finally, we ask the participants to rank on a scale from 1 to 10 for each question (i) and (ii). We did the surveys for each method \textit{separately}.

From Table \ref{tab:userstudy} and Figure \ref{fig:userstudy}, it is significant (\textit{p-value} $\leq 0.05$) that {\sc \mymethod} is able to generate more intuitive, friendly ($\mathcal{H}_1$, 6.35 v.s. 4.76 in mean ranking), and more comprehensible ($\mathcal{H}_2$, 7.35 v.s. 5.52 in mean ranking) explanation than \textsc{Lime} for general users. We also carry out an experiment that includes a visualization design showing the top 5 features and an explanation text for the top 2 features for each method as shown in Figure \ref{fig:lime}. This too results in favorable results for \textsc{\mymethod} over \textsc{Lime}.

\subsubsection{User-Study 2: How much end-users indeed understand the explanation?}
\begin{figure}[t!]
\centering
  \includegraphics[width=0.95\linewidth]{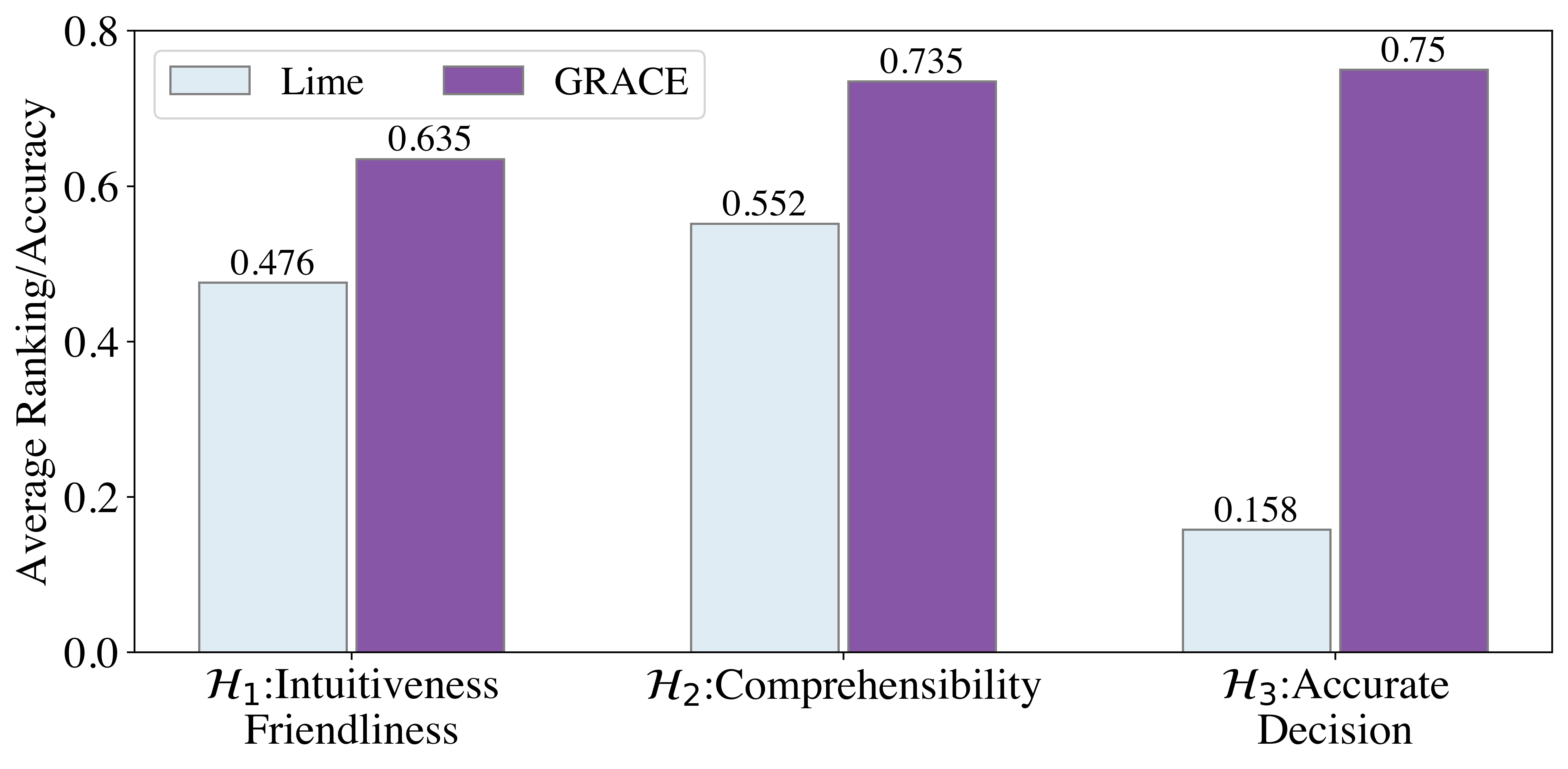}
  \caption{Comparison of generated explanation: {\sc \mymethod} v.s. {\sc Lime}. Scores are normalized to [0,1]}
  \label{fig:userstudy}
  \vskip -0.5cm
\end{figure}

In practice, ML models usually play the role of assisting human to make informative decisions \cite{lipton2018mythos}. Therefore, extending from previous experiment, we hypothesize that a good explanation should not only be comprehensible, but should also help materialize in accurate decision. Here we want to test workers' actual understanding from the explanation with alternative hypothesis ($\mathbfcal{H}_3$): users who are provided with explanation generated by \textsc{\mymethod} are better at making post-explanation decision than those provided with explanation generated by \textsc{Lime}. To test this, we first present the same prediction scenario from previous user-study, then ask each participant to analyze the explanation and adjust the sample's feature values such that the model would change its prediction (e.g., from malignant to benign). This task requires the worker to recognize from the explanation hints of both (i) what the key features are and (ii) how the changes of those features affect the model's prediction. To ensure the quality of the workers, we only select workers with ``US Graduate Degree" as a qualification provided by AMT. We use the trained model to validate the responses and report the average accuracy. From Table \ref{tab:userstudy} and Figure \ref{fig:userstudy}, it is highly significant (\textit{p-value} $\leq 0.01$) that workers who provided with explanation generated by \textsc{\mymethod} have more accurate answers than those provided with explanation generated by \textsc{Lime} ($\mathcal{H}_3$), showing 0.75 v.s 0.16 of average accuracy, respectively. In other words, explanation generated by {\mymethod} is more effective in supporting users to make tangible decisions, such as suggesting an alternative scenarios when dealing with neural network models.

\subsection{Parameter Sensitivity Analysis}
\begin{figure}[t!]
\centering
  \includegraphics[width=\linewidth]{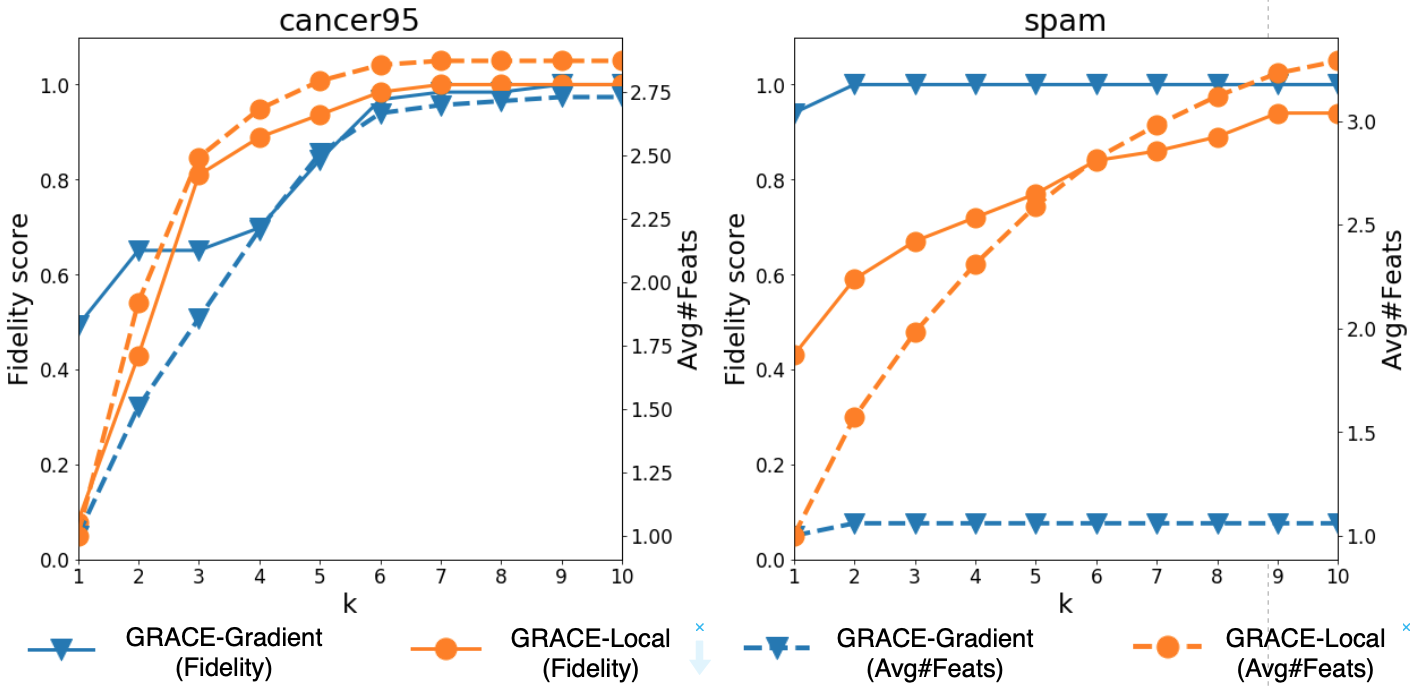}
  \caption{Effects of $K$ on $\mathbf{R}_{\mathrm{fidelity}}$ and $\mathbf{R}_{\mathrm{avg\#Feats}}$ score}
  \label{fig:ktest}
   \vskip -0.5cm
\end{figure}

\subsubsection{Effects of $K$}:
One of the important factors that largely affect the explainability of {\sc \mymethod} is the value of parameter \textit{K}, or the \textbf{maximum} number of features to change during the contrastive samples generation process. 
While a small \textit{K} is more preferable, it would become more challenging for {\sc \mymethod} to ensure perturbed samples to cross the decision boundary. This will eventually hurt \textit{$\mathbf{R}_{\mathrm{fidelity}}$}. Here we want to see how different $\mathit{K}$ values affect the generated samples' \textit{fidelity} and the number of perturbed features. For each dataset, we next train a neural network model and test this model with all values of \textit{K = \{1, 2, 3,.. 10\}} and plot it against respective $\mathbf{R}_{\mathrm{fidelity}}$ and $\mathbf{R}_{\mathrm{avg\#Feats}}$. Figure \ref{fig:ktest} reports two distinctive patterns between \textsc{\mymethod-Gradient} and \textsc{\mymethod-Local}: (i) both approaches witness gradual increment in $\mathbf{R}_{\mathrm{fidelity}}$ and $\mathbf{R}_{\mathrm{avg\#Feats}}$, with neither one of them dominates the performance (e.g., \textit{cancer95} dataset), or (ii) one of them greatly out-weights the other (e.g., \textit{spam} dataset). Overall, by increasing \textit{K}, generated samples are more faithful to the neural network model' decision boundary. Yet the average number of features needed to change to achieve so also increases, hence eventually reduce explainability.

\subsubsection{Effects of entropy threshold $\gamma$}:
Entropy threshold $\gamma$ is set to ensure that no pairs out of selected features are conveying very similar information, hence making generated samples more informative to users. Similar to the previous experiment, we keep other parameters the same while vary $\gamma$ as $\{1.0, 0.7, 0.5, 0.3\}$. The results are shown in Table \ref{tab:gamma}. We observed that $\gamma$ is not very sensitive, showing the best value of $\gamma \leq 0.5$, which can be explained that the pair of features are usually more or less predictable given the other at a specific level. However, by setting $\gamma = 0.5$, we can observe larger improvement in case of \textit{musk} and \textit{segment} dataset.

\begin{table}
    \small
    \centering
    \caption{Effects of entropy threshold $\gamma$ on $\mathbf{R}_{\mathrm{\mathrm{info-gain}}}$}
    \vskip -1em
    \begin{tabular}{lc|cccc}
    \toprule
    Dataset & Method & 1.0 & 0.7 & 0.5 & 0.3\\
    \hline
    \multirow{2}{*}{musk} & \textsc{\mymethod-Gradient} & 0.51 &  0.51 & \textbf{0.58} & \textbf{0.58}\\
             & \textsc{\mymethod-Local} & 0.36 & 0.36 & \textbf{0.54} & \textbf{0.54}\\
    \hline
    \multirow{2}{*}{segment} & \textsc{\mymethod-Gradient} & 0.57 & 0.57 & \textbf{0.59} & \textbf{0.59}\\
            & \textsc{\mymethod-Local} & 0.79 & 0.79 & \textbf{0.84} & \textbf{0.84}\\
    \bottomrule
    \end{tabular}
    \label{tab:gamma}
     \vskip -2em
\end{table}

\section{RELATED WORK}\label{sec:relatedworks}
Regarding \textit{explanation by intervention}, our Def. \ref{def:intervention} relates to \textit{Quantitative Input Influence} \cite{datta2016algorithmic}, a general framework to quantify the influence of a set of inputs on the prediction outcomes. The framework follows a two-step approach: (i) it first changes each individual feature by replacing it with a random value, and \textit{then} (ii) observes how the outcome, i.e., prediction, changes accordingly. However, we propose a more systematic way by generating a new sample at once by directly conditioning it on a contrastive outcome (\textit{X rather than Y}). A few prior works (e.g., \cite{wachter2017counterfactual,zhang2018interpreting,mao2017least}) also propose to generate contrastive samples with (i) minimal corrections from its original input by minimizing the distance: $\delta=\Vert\pmb{x} - \mathbf{\tilde{x}}\Vert_p$ and with (ii) minimal number of features needed to change to achieve such corrections. While Wachter et al. \cite{wachter2017counterfactual} use $\delta$ with $\ell_1$ norm to induce sparsity with the hope to achieve (ii), Zhang et al. \cite{zhang2018interpreting} approach the problem in a reverse fashion, in which they try to search for minimal $\delta$ w.r.t to a pre-defined number of features to be changed. Regardless, without considering the mutual information among pair-wise of features, it does not always guarantee that generated samples are informative to end-users. The work \cite{van2018contrastive} also proposes a method to use decision trees to search for a decisive threshold of feature's values at which the prediction will change, and utilize such threshold to generate explanations for neural network model' predictions. While sounds similar to our approach, this method shares a similar dis-merit with \textsc{Lime} \cite{lime} since the generated explanation is only an approximation and not faithful to the model. In this paper, we take a novel approach to generate contrastive samples that are not only contrastive but also faithful to the neural network model and  ``informative" to end-users. As regards as \textit{features selection}, we employ a forward-based approach together with  \textit{Symmetrical Uncertainty (SU)} and the approximation of features importance according to the neural network model. While there are other algorithms for ranking or selecting features (e.g., submodularity \cite{khanna2017scalable}, $\ell_1$ \cite{wachter2017counterfactual}, tree-based (\cite{van2018contrastive}, etc.), our proposed method is selected because of it is both effective (high fidelity and informative scores as a result) and easy to implement, not to mention that \textit{SU} can work with continuous features, and it also considers the bias effects in which one feature might have many diverse values than the other \cite{yu2003feature}.

\section{CONCLUSION AND FUTURE WORK}\label{sec:conclusion}
In this paper, we borrow ``contrastive explanation" and ``explanation by intervention" concepts from previous literature and develop a generative-based approach to explain neural network models' predictions. We introduce {\sc \mymethod}, a novel instance-based algorithm that provides end-users with simple natural text explaining neural network models' predictions in a contrastive \textit{``Why X rather than Y"} fashion. To facilitate such an explanation, {\sc \mymethod} extends adversarial perturbation literature with various conditions and constraints, and generates contrastive samples that are concise, informative and faithful to the neural network model's specific prediction. User-studies and quantitative experiments on several datasets of varied scales and domains have demonstrated the effectiveness of the proposed approach. There are several Interesting future directions. First, in this paper, we intervene a selected subset of features without considering conditional dependencies among all variables after such intervention. This might create undesirable samples that are unrealistic (e.g., ```a pregnant man"). Thus, we plan to address interactions among the features to generate samples that are \textit{more realistic}. Second, in this work, we assume a white-box setting that we can access the gradients of the model. We want to extend \textsc{\mymethod} for other black-box settings, gradients of which are not accessible. Since our method works exclusively for multinomial classification task, we also plan to apply it on other ML tasks such as regression, clustering, etc.
\section{ACKNOWLEDGEMENT}
This work was in part supported by NSF awards \#1742702, \#1820609, \#1909702, \#1915801 and \#1934782. We appreciate anonymous reviewers for all of their constructive comments.

\bibliographystyle{ACM-Reference-Format}
\bibliography{ref_abbr.bib}

\clearpage
\beginsupplement

\appendix

\begin{table*}[th!]
\centering
\small
\caption{The details of configuration and training parameters of neural network models on different datasets}
\vspace*{-0.4cm}
\label{tab:configuration}
\begin{tabular}{@{}c|rrrrrrrrrrr@{}}
\toprule
{Parameter} & eegeye & diabete  & cancer95 & phoneme & segment & magic & biodeg & spam & cancer92 & mfeat & musk \\ 
\hline
Size of Hidden Layers & [40,30] & [15, 7] & [15, 15] & [20, 5] & [30, 10] & [35, 20] & [60, 50] & [50, 30] & [50, 20] & [100, 50] & [100, 100]\\
Batch Size & 512 & 512 & 512 & 512 & 512 & 512 & 512 & 512 & 512 & 512 & 512\\
Learning Rate & 0.01 & 0.01 & 0.001 & 0.001 & 0.01 & 0.001 & 0.01 & 0.001 & 0.001 & 0.001 & 0.0001\\
Early-Stopping Patience & 5 & 3 & 3 & 3 & 3 & 4 & 3 & 3 & 3 & 5 & 3\\
Maximum Epochs & 500 & 500 & 500 & 500 & 500 & 500 & 500 & 500 & 500 & 500 & 500\\
\bottomrule
\end{tabular}
\vspace*{-0.3cm}
\end{table*}

\begin{table*}[th]
\centering
\small
\caption{Prediction scenario from \textit{cancer95} dataset used in case-study and user-studies }
\vspace*{-0.3cm}
\label{tab:casestudy:example}
\begin{tabular}{l|cccccccccc}
\toprule
Feature & Bare\_Nuclei & BlaChr & MarAdh & CluThic & Mitoses & CelSizUni & CelShaUni & NorNuc & SinEpCeSi & \textbf{Model Prediction}\\
\hline
Patient & 10.0 & 4.0 & 5.0 & 8.0 & 1.0 & 5.0 & 5.0 & 3.0 & 2.0 & \textbf{88\% Malignant}\\
\bottomrule
\end{tabular}
\vspace*{-0.3cm}
\end{table*}

\begin{table}[h]
\centering
\small
\caption{The details of parameters of {\sc \mymethod}}
\vspace*{-0.3cm}
\label{tab:config:grcs}
\begin{tabular}{|l|c|}
\toprule
Parameter & Value\\ 
\hline
$\mathit{K}$, $\gamma$, STEPS & 5, 0.5, 200\\
\hline
Nearest Neighbor parameter (n\_neighbors) & 4\\
\bottomrule
\end{tabular}
\end{table}
\vspace*{-3mm}

\section{APPENDIX ON REPRODUCIBILITY}
In this section, we provide the details of experimental configuration and the designs of our user-studies to facilitate the reproducibility of our work.

\subsection{Source Code, Software, and Dataset}
Software wise, we use \textit{Python} (version 3.7.3) as the main programming language, \textit{Scikit-learn} (version 0.21.3) and \textit{PyTorch} (version 1.4.0) as the main machine learning frameworks. All eleven datasets used in this paper are publicly available in the UCI Machine Learning Repository \cite{uci}.

\subsection{Handling of Categorical Features}
\textit{(Updated October 26)} The current algorithm can handle categorical features with two steps: (i) encoding all categorical features using a numerical embeddings layer and training this layer with the rest of the model, (ii) projecting the perturbed embedding values during the generation step (Sec. \ref{sec:generation}) back into the discrete space using a \textit{nearest neighbor} searching algorithm on the learned embedding space.

\subsection{Evaluation of Generated Samples}
In Section \ref{sec:experiment}, we compare {\sc \mymethod} with {\sc DeepFool} and {\sc NearestCT} baseline on the quality of generated contrastive samples to explain predictions of neural network model $\mathit{f}(\cdot)$ with parameters and configuration as follows.

\subsubsection{Neural Network Model $\mathit{f}(\cdot)$}\label{sec:neural_models}
We employ two hidden layers fully-connected-networks with \textit{ReLu} activation function, followed by a softmax layer to train a neural network model for each dataset with parameters and configuration described in Table \ref{tab:configuration}. We initialize all the weight tensors and biases of the model using \textit{Glorot initialization} \cite{glorot2010understanding} and value of 0.01, respectively. We use \textit{Adam} optimizer \cite{kingma2014adam} with $\beta_1=0.9$, $\beta_2=0.999$, and $\epsilon=1e-8$ to train the model with \textit{Cross Entropy Loss}.

\subsubsection{{\sc \mymethod}}
We follow the generation algorithm as described in Section \ref{sec:generation}, particularly Alg. \ref{alg:grcs}, Alg. \ref{alg:generation}, and Alg. \ref{alg:grcs:entropy}. Implementation of \textit{Symmetrical Uncertainty (SU)} function is retrieved from public repository for \textit{Fast Correlation-Based Filter Feature Selection}: \url{https://github.com/shiralkarprashant/FCBF}. We use Logistic Regression as the explainable ML classifier $\mathit{g}(\cdot)$ used in $\mathit{W}_{Gradient}$ function described in Section \ref{sec:generation}. We set all parameters following Table. \ref{tab:config:grcs}. The descriptions of major parameters are as follows.

\begin{enumerate}
    \item $K$: Maximum number of features to perturb (as in Alg. \ref{alg:grcs})
    \item $\gamma$: Entropy upper-bound threshold (as in Alg. \ref{alg:grcs:entropy})
    \item STEPS: Maximum number of steps to project $\tilde{\pmb{x}}$ on the contrastive class:  (as in Alg. \ref{alg:generation})
    
\end{enumerate}

\subsubsection{{\sc DeepFool}}
Since {\sc DeepFool} is originally developed for image data, we adapt a publicly available repository  at \url{https://github.com/LTS4/DeepFool} to our tabular datasets.

\subsection{Evaluation of Explanation}
We compare our proposed framework, {\sc \mymethod}, with {\sc Lime} on the quality and effectiveness of generated explanation. We randomly select a sample from \textit{cancer95} dataset and study its prediction throughout Section \ref{sec:case_study}. Its feature values and prediction is shown in Table. \ref{tab:casestudy:example}.

\subsubsection{Case-study: breast-cancer diagnosis}
\begin{figure*}[th]
  \centering
  \includegraphics[width=\linewidth]{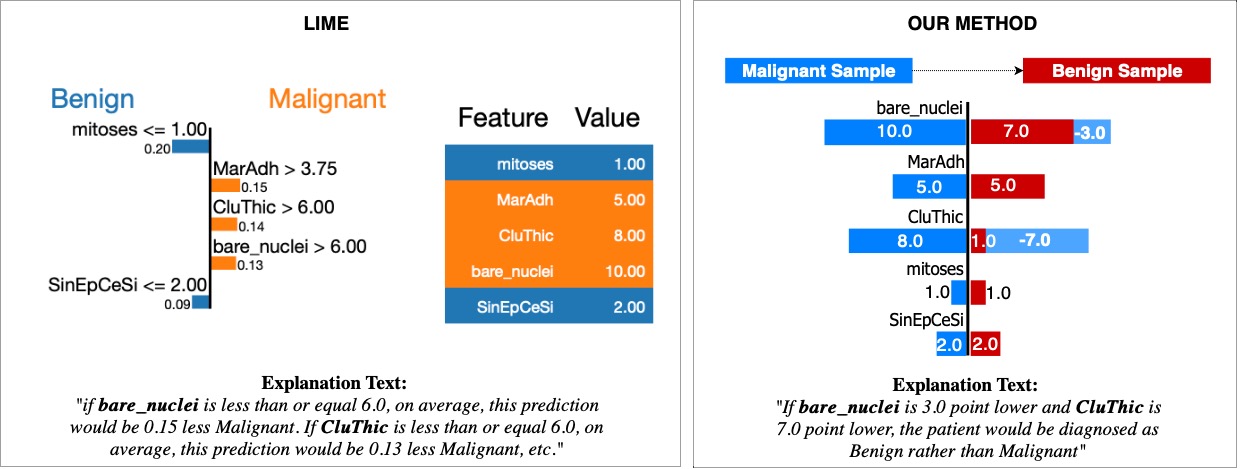}
  \caption{Example explanation produced by {\sc Lime} and our method (\textsc{\mymethod}).}
  \label{fig:lime}
\end{figure*}

:
We study the prediction scenario of patient described in Table. \ref{tab:casestudy:example}. Figure \ref{fig:lime} shows the explanation generated by {\sc Lime} and {\sc \mymethod} for the scenario. Even though {\sc \mymethod} focuses on generating explanation in natural text and visualization is not our main contribution, we attach a potential visualization design corresponding to the generated explanation text as seen in Figure \ref{fig:lime}.

\subsubsection{User-Study 1: Intuitiveness, friendliness \& comprehensibility}
\begin{figure*}[th]
  \centering
  \includegraphics[width=0.85\linewidth]{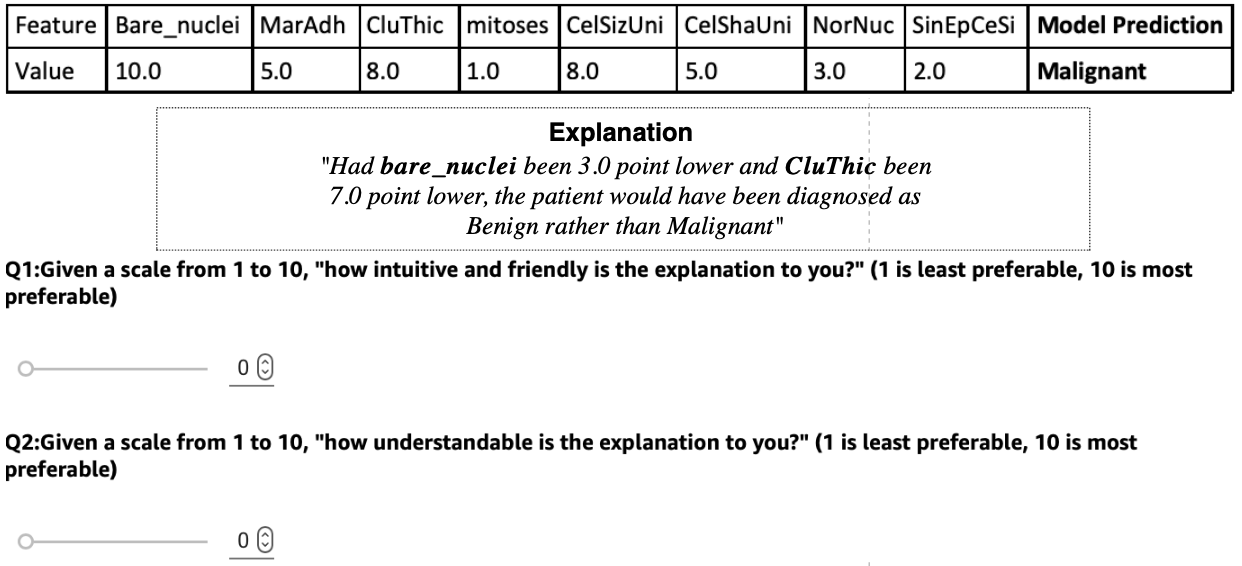}
  \caption{Interface of User-study 1 ({\sc \mymethod}: Intuitiveness, friendliness \& comprehensibility).}
  \label{fig:userstudy1:grace}
\end{figure*}

:
We first present the AMT workers the prediction scenario as in Table. \ref{tab:casestudy:example}. To ensure the quality of the workers, we only recruit workers with the approval rate of assignments greater than 95\%. To ensure the quality of the responses, we filter out ones that spend less than 1.5 minutes. Eventually, we have 37 workers participated with a total of 44 valid responses, 23 and 21 of which are asked to assess {\sc \mymethod}'s, and {\sc Lime}'s explanation, respectively. An example of the task interface is depicted in Figure \ref{fig:userstudy1:grace}.

\subsubsection{User-Study 2: How much end-users indeed understand the explanation?}
\begin{figure*}[th]
  \centering
  \includegraphics[width=0.8\linewidth]{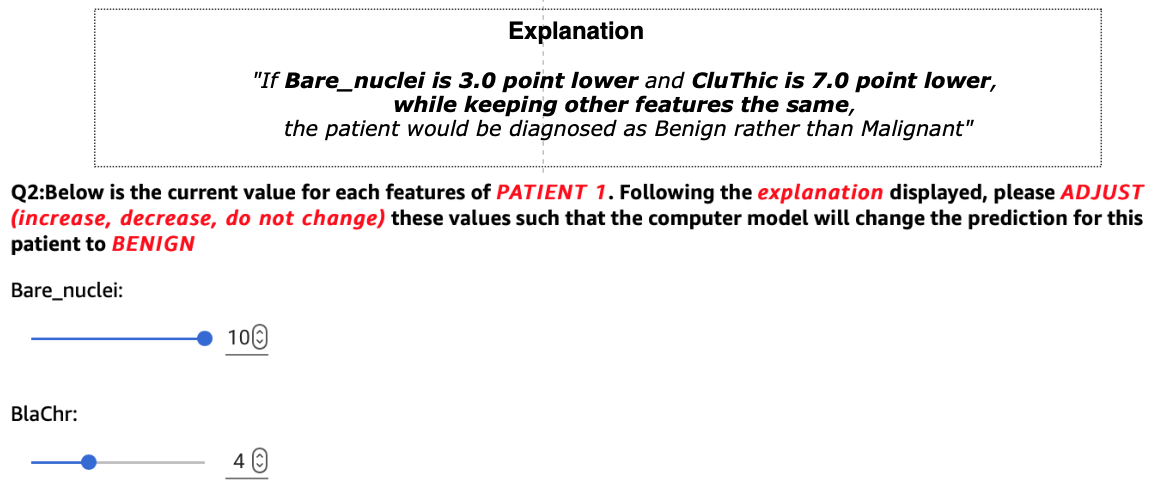}
  \caption{Interface of User-study 2 ({\sc \mymethod}: How much end-users indeed understand the explanation?) Note that input controls for other features (rather than \textit{Bare\_nuclei}, \textit{BlaChr}) and the feature of the sample is omitted in the figure due to space issue.}
  \label{fig:userstudy2:grace}
\end{figure*}

We also present the AMT workers the prediction scenario as in Table. \ref{tab:casestudy:example}. To ensure the quality of the workers, we apply two recruitment requirements provided by AMT: (i) the approval rate of assignments of any workers must be greater than 95\%, (ii) the workers must have a "US Graduate Degree". To ensure the quality of the responses, we filter out ones that spend less than 3 minutes. Eventually, we have 24 workers participated with a total of 39 valid responses, 20 and 19 of which are provided with {\sc \mymethod}'s, and {\sc Lime}'s explanation, respectively. An example of the task interface for {\sc \mymethod} is depicted in Figure \ref{fig:userstudy2:grace}.

\end{document}